\documentclass[iicol,pdflatex]{sn-jnl}

\usepackage[subpreambles=true]{standalone}
\usepackage{geometry}
\usepackage{float}
\usepackage[dvipsnames]{xcolor}
\usepackage{graphicx}
\usepackage{manyfoot}
\usepackage{amsmath}
\usepackage{amsfonts}
\usepackage{amssymb}
\usepackage{multirow}
\usepackage{multicol}
\usepackage{csquotes}
\usepackage{lmodern}
\usepackage{makecell}
\usepackage{booktabs, hhline}
\usepackage[misc]{ifsym}
\usepackage{subfiles}

\usepackage{fancyhdr}

\usepackage{silence}
\usepackage{caption}
\usepackage{subcaption}

\usepackage[style=authoryear,sorting=nty,eprint=false,doi=false,url=true,maxbibnames=30,mincitenames=1,maxcitenames=2]{biblatex} 
\usepackage{algorithm}
\usepackage[noend]{algpseudocode}

\DeclareCaptionLabelFormat{tablelabelformat}{{\small Tab.\space \arabic{table}:}}
\DeclareCaptionLabelFormat{figurelabelformat}{{\small Fig.\space \arabic{figure}:}}
\DeclareCaptionLabelFormat{algorithmlabelformat}{{\small Alg.\space \arabic{algorithm}:}}
\DeclareCaptionLabelSeparator{captionlabelsep}{\space}
\DeclareCaptionTextFormat{captiontextformat}{{\small #1}}
\makeatletter
  \patchcmd{\thebibliography}{\leftmargin\labelwidth}{\leftmargin0pt}{}{}
  \patchcmd{\thebibliography}{\labelsep}{0pt}{}{}
  \patchcmd{\@bibitem}{\item}{\item[]}{}{}
\makeatletter

\algrenewcommand\algorithmicindent{.9em}
\algnewcommand\Part[1]{\item[\textbf{#1:}]}
\algdef{S}[FOR]{ForUnroll}[1]{\algorithmicfor\ #1\ \textbf{unroll}}
\algnewcommand\Proc[2]{\item[\textbf{Procedure:}] \Call{#1}{#2}}
\algnewcommand\Label[1]{\item[\texttt{#1}:] }

\newcommand{\algstrut}[1][\algruledefaultfactor]{\vrule width 0pt depth .25\baselineskip height #1\baselineskip\relax}
\newcommand*{\algrule}[1][\algorithmicindent]{\hspace*{.5em}{\color{lightgray}\vrule}\algstrut\hspace*{\dimexpr#1-.5em}}
\newcommand{\alghline}[1][.2pt]{\par\vskip.5\baselineskip\hrule height #1\par\vskip.5\baselineskip}

\renewcommand{\thetable}{\arabic{table}}

\newcommand{\len}[1]{\vert#1\vert}

\makeatletter
\newcount\ALG@printindent@tempcnta
\def\ALG@printindent{%
    \ifnum \theALG@nested>0
    \ifx\ALG@text\ALG@x@notext
    \else
    \unskip
    \ALG@printindent@tempcnta=1
    \loop
    \algrule[\csname ALG@ind@\the\ALG@printindent@tempcnta\endcsname]%
    \advance \ALG@printindent@tempcnta 1
    \ifnum \ALG@printindent@tempcnta<\numexpr\theALG@nested+1\relax
    \repeat
    \fi
    \fi
}%

\patchcmd{\ALG@doentity}{\noindent\hskip\ALG@tlm}{\ALG@printindent}{}{\errmessage{failed to patch}}

\AtBeginEnvironment{algorithmic}{\lineskip0pt}

\begin{document}

\title[FIMD: Fast Isolated Marker Detection for UV-Based Visual Relative Localisation in Agile UAV Swarms]
{{FIMD:} Fast {Isolated} Marker Detection for UV-Based Visual Relative Localisation in Agile UAV Swarms}

\author*[1]{\fnm{Vojtěch} \sur{Vrba}} \email{vojtech.vrba@fel.cvut.cz}
\author[1]{\fnm{Viktor} \sur{Walter}} \email{viktor.walter@fel.cvut.cz}
\author[1]{\fnm{Petr} \sur{Štěpán}} \email{petr.stepan@fel.cvut.cz}
\author[1]{\fnm{Martin} \sur{Saska}} \email{martin.saska@fel.cvut.cz}

\affil*[1]{\orgdiv{Multi-Robot Systems Group, Department of Cybernetics}, \orgname{Faculty of Electrical Engineering, Czech Technical University in Prague}, \orgaddress{\street{Technicka 2}, \postcode{16000} \city{Praha}, \country{Czech Republic}}}

\abstract{
A novel approach for the fast onboard detection of isolated markers for visual relative localisation of multiple teammates in agile UAV swarms is introduced in this paper. As the detection forms a key component of real-time localisation systems, a three-fold innovation is presented, consisting of an optimised procedure for CPUs, a GPU shader program, and a functionally equivalent FPGA streaming architecture. For the proposed CPU and GPU solutions, the mean processing time per pixel of input camera frames was accelerated by two to three orders of magnitude compared to the {unoptimised state-of-the-art approach}. For the localisation task, the proposed FPGA architecture offered the most significant overall acceleration by minimising the total delay from camera exposure to detection results. Additionally, the proposed solutions were evaluated on various 32-bit and 64-bit embedded platforms to demonstrate their efficiency, as well as their feasibility for applications using low-end UAVs and MAVs. Thus, it has become a crucial enabling technology for agile UAV swarming.
}


\maketitle

\section{Introduction}
A robust and accurate localisation system is a key component for the close interaction and collaboration of Unmanned Aerial Vehicles (UAVs). Such a system is an essential tool for the fulfilment of various related objectives, such as mutual collision avoidance, swarm cohesion and alignment, and multi-agent trajectory planning \parencite{Yasin2020}.

\begin{figure}[t]
  \centering
  \includegraphics[width=\linewidth]{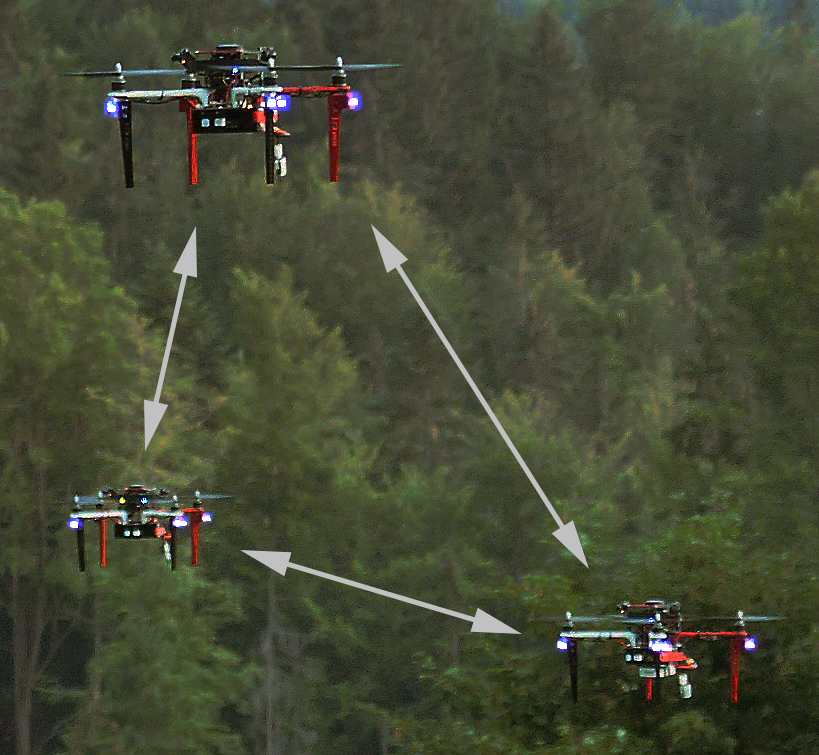}
  \caption{Real-world application of the UVDAR system.}
  \label{fig:photo_swarm}
\end{figure}

The commonly used localisation systems are dependent on external infrastructure, including motion capture (MoCap) systems \parencite{Abdelkader2021} used for well defined indoor environments, and global navigation satellite systems (GNSS) used for outdoor, open-sky conditions with a direct satellite link \parencite{Lopez2021}. In general, these systems provide an absolute estimate of the poses and velocities of the swarm agents given the corresponding frame of reference. Despite the high precision these methods may provide under ideal conditions, real-world applications usually have to deal with GNSS-denied, obstacle-dense, or unknown environments. Such conditions lead to the need for localisation methods based on relative inter-agent measurements. It was shown that distance-based methods utilising omnidirectional radio-based communication standards, such as Wi-Fi \parencite{Kealy2019}, Bluetooth \parencite{Coppola2018}, or ultra-wideband radio (UWB) \parencite{Goel2017}, are able to provide accurate estimates of relative agent poses. {When compared to the former two standards, the UWB-based approach offers the most accurate distance measurements between network nodes. Consequently, recent literature has proposed its fusion with other sensors, such as stereo cameras \parencite{Xu2022} or LIDARs \parencite{Queralta2022}.} However, with an increasing number of swarm agents, the probability of depleting the network resources increases. This can possibly result in signal jamming and an overall decrease of the localisation system robustness when approaching wireless network limits. Another category of relative localisation techniques relevant to this work is based on computer vision. This modality offers better scalability and saturation immunity while introducing new challenges.

{In general, vision-based localisation methods may be either marker-less or they may use additional markers attached to the targets. The former methods perform recognition of individual targets directly in camera frames, usually by employing convolutional neural networks (CNNs) to handle detection of miscellaneous UAV projections \parencite{Vrba2019, Schilling2021}. However, the CNNs must be trained using well-prepared datasets before the actual real-world deployment of the UAVs. The latter methods are based on detection of predefined markers which may be either passive or active. Methods utilising passive markers}, such as \parencite{Krajnik2014}, suffer primarily from dependency on strong lighting conditions, which prevents their general deployment in diverse environments. {Methods based on active markers} eliminate this dependency by utilising light-emitting diodes (LEDs) working either in the visible spectrum \parencite{Dias2016, Stuckey2023} or in the infra-red (IR) spectrum \parencite{Cutler2013, Faessler2014, Xun2023}. However, these methods are usually applicable in indoor environments only due to an ubiquitous amount of sunlight outdoors. For outdoor environments, it was shown in \parencite{Walter2018_1} that the usage of ultra-violet (UV) light is able to overcome this challenge.

\begin{figure}[t!]%
\centering
\begin{minipage}[b]{.59\linewidth}
\fontsize{8}{11}\selectfont
  \centering
  \includegraphics[width=\linewidth]{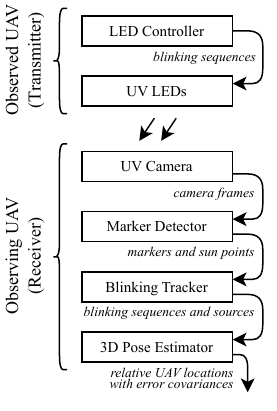}
  \caption{The UVDAR system pipeline.}
  \label{fig:UVDAR_System_Pipeline}
\end{minipage}%
\begin{minipage}[b]{.36\linewidth}
\fontsize{8}{11}\selectfont
\vspace{0.5em}
  \centering
    \begin{tabular}{@{}c@{}}
    \includegraphics[width=\linewidth]{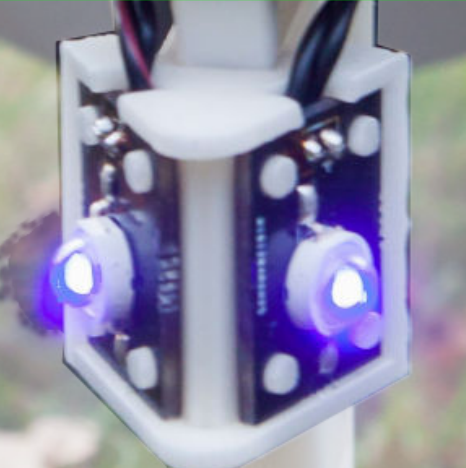} \\ {(a) UV power LEDs} \vspace{0.1cm} \\
    \includegraphics[width=\linewidth]{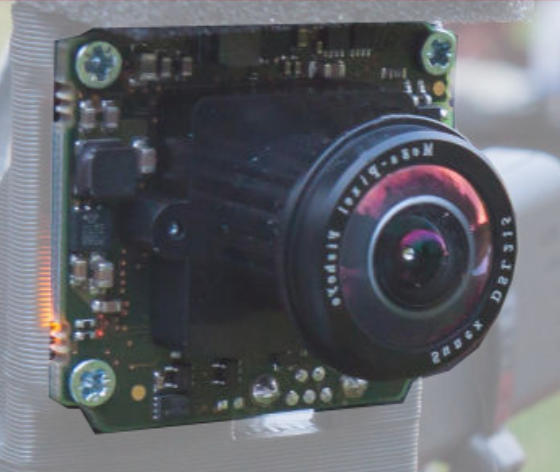} \\ {(b) UV-sensitive camera} \\
    \end{tabular}
  \caption{Components of the UVDAR system.}
  \label{fig:UV_components}
\end{minipage}
\end{figure}

The Ultra-Violet Direction And Ranging (UVDAR) system \parencite{Walter2018_1, Walter2018_2, Walter2019} is a vision-based distributed relative localisation framework. The system, illustrated in Fig. \ref{fig:photo_swarm}, uses UV LEDs as UAV markers together with UV bandpass filters mounted on fisheye camera lenses. Additionally, the UV LEDs of different UAVs emit predefined unique blinking sequences that are used for their identification, as well as for the estimation of their relative poses and flight trajectories. The system consists of a two-part pipeline, as shown in Fig. \ref{fig:UVDAR_System_Pipeline}. The transmitter emits the blinking sequences via an optical channel using UV LEDs (Fig. \ref{fig:UV_components}a). The receiver consists of four essential elements: a monochromatic camera with UV bandpass filters and a fisheye lens (Fig. \ref{fig:UV_components}b), a detector of isolated points in camera frames, a tracker of the blinking sequences between subsequent frames, and a pose estimator used for computation of the locations and orientations of the observed UAVs relative to the observing UAV.

The UVDAR system forms an essential localisation tool used by teams of closely collaborating UAV agents. All the necessary computer vision calculations are performed on the onboard computers of the individual UAV agents, while consuming a significant amount of their computational resources. Moreover, with the increasing agility of the UAVs, these requirements tend to grow rapidly while reflecting the requisite increase of the sensors' frame rate.

In this paper, the theoretical background of an {unoptimised} detection algorithm for the isolated LED markers of the UVDAR system is investigated. Next, its optimised CPU-based variant is presented as a pseudocode, which is easily interpreted as a small set of assembly instructions. Its parallel variants are also presented, i.e., a functionally equivalent GPU shader program and an FPGA streaming architecture. Lastly, the performance of the presented solutions is evaluated along with a state-of-the-art approach on an onboard computer, as well as several other CPU+GPU boards. In the case of the FPGA solution, the influence of the camera interface on the overall performance is also discussed.

\subsection{Problem statement}

This paper tackles the problem of fast real-time visual localisation for agile multi-robot systems, based solely on onboard sensors without any radio communication support. In particular, the main focus is on the insufficient performance of a state-of-the-art computer vision algorithm for the detection of isolated LED markers when high frame rate camera sensors are utilised under the limitations of the onboard computational resources of UAVs.

\subsection{Contribution}

A novel three-fold approach to the fast detection of isolated UV LED markers in camera frames for onboard relative localisation of agile team members in multi-robot systems is introduced. First, this paper presents a sequential CPU-based approach to the detection task based on minimising the required number of processor instructions per each pixel of the camera frame, and on an appropriate assembly procedure branching. Second, a parallel GPU-based approach via an OpenGL ES compute shader is presented, utilising integrated graphics cards available in the majority of modern CPU chipsets for desktop and embedded systems. Third, a functionally equivalent FPGA architecture operating directly on a camera pixel stream is proposed. A real-world dataset is used to compare all the proposed solutions with the state-of-the-art UVDAR approach on various platforms. Apart from significant acceleration gained by the proposed CPU and GPU solutions, it is shown that the FPGA approach provides the most perceptible enhancement of the detection part of the localisation system.

\section{Preliminaries}

{Multiple approaches to the detection of blobs of general size and shape in greyscale images have been presented in the literature \parencite{Damerval2007, Petrovic2017, AcevedoAvila2016}. However, much larger computational resources are required by these general approaches, and they cannot easily be scaled down for the detection task presented in this work. In contrast, the marker detection algorithms described in this work are based mostly on integer arithmetic and a simple decision logic. Therefore, the derived solutions can be easily implemented for low-power CPUs, embedded GPUs and low-end FPGAs.}

\begin{figure*}[t]
\fontfamily{ptm}\fontsize{7.5}{10}\selectfont
  \centering
  \includegraphics[width=\linewidth]{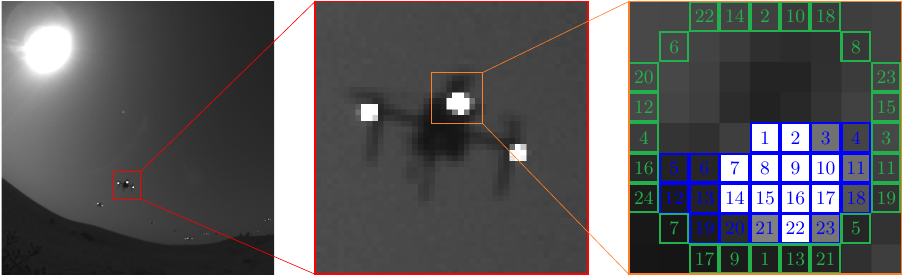}
    \begin{tabular}{p{0.29\linewidth}@{\hspace{0.05\linewidth}}p{0.29\linewidth}@{\hspace{0.06\linewidth}}p{0.29\linewidth}}
    {(a) Fisheye camera frame in a narrow band of the UV spectrum containing a projection of the sun and multiple UAV agents.} & 
    {(b) Detail of a single UAV with 3 distinguishable isolated UV LED markers to be detected.} & 
    {(c) Detail of a marker with a denoted pixel {evaluation} order for the segment test (green) and for the interior peak search (blue).}
    \end{tabular}
  \caption{Illustration of the algorithm for the detection of the isolated LED markers.}
  \label{fig:detector_example}
\end{figure*}

The algorithm for the detection of the isolated LED markers in UV-based localisation system \parencite{Walter2018_1} is derived from the original Features from Accelerated Segment Test (FAST) algorithm \parencite{Rosten2005}. In this algorithm, if a pixel's intensity differs greatly from enough pixels in the surrounding circular neighbourhood of the pixel (i.e., the pixel passes the segment test), then it can be considered as a corner. In the discussed use case, the projections of the mutually distant LED diodes are visually separated from each other with a perceptible contrast to the background in the camera images. Therefore, it is desirable to detect pixels that differ from all pixels in a neighbourhood by a pre-defined intensity threshold {\parencite{Xiong2013}}. In addition, it is desirable to detect pixels projecting from the sun and its reflections (called \emph{sun points} later on), which saturate much of the images. Hence, an inverse condition is used to compare the intensity of pixels with other pixels in the neighbourhood to detect those with similar brightness. LED markers in the vicinity of the sun points may then be eliminated to improve the robustness against false positives caused by glare and image noise.

\begin{algorithm}[t!]
\fontfamily{ptm}\fontsize{8.5}{10.5}\selectfont
\caption{{Unoptimised isolated marker detection.}}
\label{alg:unoptimised}
\begin{algorithmic}[1]
\Part{Constants}
    \Statex Image width $W$,
    \Statex Image height $H$,
    \Statex Threshold for marker test $T_{m}$,
    \Statex Threshold for sun point test $T_{s}$,
    \Statex Differential threshold $T_{d}$,
    \Statex Array of 2D relative coordinates $N_{\rho,\text{b}}$ of boundary pixels of the circular neighbourhood with a radius $\rho \in \{3,4\}$,
    \Statex Array of 2D relative coordinates $N_{\rho,\text{i}}$ of interior pixels of the circular neighbourhood
\Part{Variables}
    \Statex Camera image array $I[H, W]$, \Comment{2D, row-major order}
    \Statex Array of detected markers $M$,
    \Statex Array of detected sun points $S$,
    \Statex Auxiliary 2D boolean array $A$
\alghline
\Proc{{UIMD}MarkerDetection}{$I$}
    \State $M \gets \varnothing$, $S \gets \varnothing$, $A \gets \{\text{False}\}$ 
    \For{\textbf{each} row $r \in [1, ..., H]$}
        \For{\textbf{each} column $c \in [1, ..., W]$}
            \If{$A[r, c] = \text{False}$}
                \If{$I[r, c] > T_m$}
                    \State $p_m$, $p_s$ $\gets$ \Call{{UIMD}SegmentTest}{$I$, $r$, $c$}
                    \If{$p_m = \text{True}$}
                        \State $r_p$, $c_p$ $\gets$ \Call{{UIMD}PeakSearch}{$I$, $A$, $r$, $c$}
                        \State $M \gets M + [(r_p, c_p)]$
                    \ElsIf {$p_s = \text{True}$}
                        \If{$c_s = \len{{N_{\rho,\text{b}}}}$}
                            \State $S \gets S + [(r, c)]$
                        \EndIf
                    \EndIf
                \EndIf
            \EndIf
        \EndFor
    \EndFor
    \State \Return $M$, $S$
\end{algorithmic}
\end{algorithm}

 {The approach inspired by the FAST algorithm that is used by the solution in \parencite{Walter2018_1} was named Unoptimised Isolated Marker Detection (UIMD)} and is illustrated in Fig. \ref{fig:detector_example} and stated in Alg. \ref{alg:unoptimised}. This {unoptimised} approach is used later on in Sec. \ref{sec:experiments} for comparison with the novel approach.

Similar to the original FAST algorithm, the Alg. \ref{alg:unoptimised} is based on iteration over all pixels (i.e., pixel intensity values) of the camera image and uses Bresenham's circle approximation \parencite{Bresenham1977} for the arrays $N_{\rho,\text{b}}$ (the boundary array) and $N_{\rho,\text{i}}$ (the interior array). However, any approximation of a circle (or any closed shape in general) may be used for the segment test and the interior peak search. Unlike the original FAST algorithm which searches for contiguous sets of pixels on the boundary meeting the brightness criteria, the Alg. \ref{alg:unoptimised} requires all the boundary pixels to pass the segment test. Thanks to the locally concave distribution of pixel brightness associated with a single LED projection, a straightforward acceleration is possible by testing the most mutually distant pixels first. {This introduces a specific pixel evaluation order, shown as the green positions in Fig. \ref{fig:detector_example}c, where the numbers indicate the sequence in which the algorithm examines each pixel.}

\begin{algorithm}[h]
 \fontfamily{ptm}\fontsize{8.5}{10.5}\selectfont
\caption{{Unoptimised segment test.}}
\label{alg:unoptimised_segment_test}
\begin{algorithmic}[1]
\Proc{{UIMD}SegmentTest}{$I$, $r$, $c$}
    \State $p_m \gets \text{False}$ \Comment{Potential of being a marker (boolean)}
    \State $p_s \gets \text{False}$ \Comment{Potential of being a sun point (boolean)}
    \If{$I[r, c] > T_s$}
        \State $p_s \gets \text{True}$ \Comment{Sun point must be brighter than $T_s$}
    \EndIf
    \For{\textbf{each} radius $\rho \in [3, 4]$}
        \State $c_s \gets 0$ \Comment{Count of successful sun point tests}
        \State $p_m \gets \text{True}$ \Comment{Initially assume the pixel is a marker}
        \For{\textbf{each} $(y, x) \in N_{\rho,\text{b}}$}
            \If{{\textbf{not} ($0 \leq r+y < H$ \textbf{and} $0 \leq c+x < W$)}} 
                \State $p_m \gets \text{False}$ \Comment{The point must lie inside the image}
                \State \textbf{break}
            \EndIf
            \If{$I[r, c] - I[r+y, c+x] < T_d$}
                \State $p_m \gets \text{False}$  \Comment{Bright neighbourhood, not a marker}
                \If{$p_s = \text{False}$}
                    \textbf{break} \Comment{Not a sun point either}
                \Else
                    \; $c_s \gets c_s + 1$ \Comment{Still can be a sun point}
                \EndIf
            \Else
                \; $p_s \gets \text{False}$ \Comment{{Dark neighbourhood, not a sun pt.}}
            \EndIf
        \EndFor
        \If{$p_m = \text{True}$}
            \textbf{break} \Comment{Already a marker, quit}
        \EndIf
    \EndFor
    \State \Return $p_m$, $p_s$
    
\end{algorithmic}
\end{algorithm}

\begin{algorithm}[t]
 \fontfamily{ptm}\fontsize{8.5}{10.5}\selectfont
\caption{{Unoptimised interior peak search.}}
\label{alg:unoptimised_peak_search}
\begin{algorithmic}[1]
\Proc{{UIMD}PeakSearch}{$I$, $A$, $r$, $c$}
    \State $r_p, c_p \gets r, c$ \Comment{Peak pixel coordinates}
    \State $p_i \gets 0$ \Comment{Peak pixel intensity value}
    \For{\textbf{each} $(y, x) \in N_{4,\text{i}}$}
        
        \If{{\textbf{not} ($0 \leq r+y < H$ \textbf{and} $0 \leq c+x < W$)}} 
            \State \textbf{break} \Comment{The point must lie inside the image}
        \EndIf
        \If{$A[r+y, c+x] = \text{False}$}
            \If{$I[r+y, c+x] > p_i$}
                \State $p_i \gets I[r+y, c+x]$
                \State $r_p, c_p \gets r+y, c+x$
            \EndIf
            \State $A[r+y, c+x] \gets \text{True}$
        \EndIf
    \EndFor
    \State \Return $r_p$, $c_p$
\end{algorithmic}
\end{algorithm}

In addition, the {unoptimised} approach utilises two acceleration methods based on the most common distribution of pixel intensities belonging to a LED marker. The first method is a peak search in the neighbourhood interior, i.e., coordinates of the pixel with the peak intensity inside the tested circular neighbourhood are selected for the final marker. The second method is clearing the interior, i.e., boolean values of the auxiliary array $A$ corresponding to the coordinates inside the interior of the neighbourhood are set in order to avoid the repeated detection of the markers. It is sufficient to search and clear the lower half of the interior (i.e., the blue pixel positions in Fig. \ref{fig:detector_example}c), because the upper half has already been processed in the previous iterations of the row-column loop in the Alg. \ref{alg:unoptimised}.

The segment test is performed at first for a smaller circle radius $\rho = 3$, and then repeated if the detection was not successful for the larger radius $\rho = 4$, as shown in Alg. \ref{alg:unoptimised_segment_test}. {It should be noted that the particular choice of the radii is based on data from real-world experiments and is discussed later on in Sec. \ref{sec:experiments}.}

Evidently, the choice of values for all three thresholds $T_m$, $T_s$, and $T_d$ has a great impact on the performance of the algorithm when used with a CPU, as they determine the number of numeric comparisons performed for each of the image pixels. If a pixel has maintained the potential to be a marker, then the search for a peak intensity inside the interior of the larger circular neighbourhood is performed, as shown in Alg. \ref{alg:unoptimised_peak_search}.

In practice, the detected markers are eventually filtered out based on their L2-distances from the detected sun points using a straightforward point-to-point filtering algorithm with $\mathcal{O}(\len{M} \cdot \len{S})$ complexity. The reasoning behind the filtering is that the sun projection may be partially occluded (e.g., when the sun is located behind trees w.r.t. the camera), {typically causing a glare around silhouettes, which leads to the detection of false isolated markers. Furthermore, false detection results may stem from material imperfections and internal reflections within the lens, leading to artifacts such as lens flare \parencite{Vitoria2018}}.

\section{Methodology}

{The proposed approach aims to design the Fast Isolated Marker Detection (FIMD) that is required for agile multi-robot aerial operations. The FIMD approach includes a sequential CPU-based solution, a parallel GPU-based solution and a streaming FPGA architecture. This combination of solutions ensures that the FIMD approach is versatile and can be adapted to various hardware configurations and operational requirements.}

\subsection{{FIMD-CPU:} Optimised sequential solution for a CPU}

{The objective is to design an approach applicable to small microcontrollers (MCUs) to enable swarms of micro-scale aerial vehicles, as is required by most applications.}

In order to minimise the number of processor instructions needed to process each image pixel, the proposed approach eliminates most of the variables used to store intermediate results. Consequently, this approach also avoids all unnecessary stack and RAM data transfers and algorithmic computations by taking advantage of the processor branching mechanism applied in unrolled loops from the {unoptimised} sequential procedures. That in turn makes the marker detection procedure suitable for low-end 32-bit microcontrollers.

The proposed solution is described in detail in Alg. \ref{alg:cpu}. The procedure is based on pixel-by-pixel processing in the default row-major order using hard-coded 1D linear relative address offsets, which are dependent on the image width $W$. The absolute values of all neighbourhood offsets are smaller than the central pixel offset from the first segment pixel, which is also the first pixel processed by the procedure. Lastly, a termination pixel sequence $E$ is written to the end of the image and its presence is always checked at the central pixel offset distance (line 4), thus an additional termination variable is unnecessary. Two bytes are suggested for the termination pixel sequence $E$, namely \texttt{0x00} and \texttt{0xFF}, as the presence of two oppositely saturated pixels next to each other is improbable due to the combined properties of the camera imaging sensor, the UV bandpass filters, and the fisheye lens used by the UVDAR system, which is supported by real-world experiments.

The most computationally expensive part begins at the label \texttt{LOOP}, as it is performed for every pixel in the image until the termination pixel sequence $E$ is found (lines 4-6). The mutual exclusivity of the marker and sun point pixel potentials enables the procedure to evaluate only the appropriate potential later by testing the first pixel of the circular neighbourhood boundary (lines 7-10).

\begin{algorithm}[H]
 \fontfamily{ptm}\fontsize{8.5}{10.5}\selectfont
\caption{Optimised sequential solution for a CPU.}
\label{alg:cpu}
\begin{algorithmic}[1]
\Part{Constants} \Comment{Header-defined for pre-processor}
    \Statex Relative 1D offset of the central pixel $O_c \gets (W+1)\cdot \rho$,
    \Statex Relative 1D offsets for the boundary $r_{\text{b},k} \gets Wy+x$ for each ($k$-th) point $(y, x) \in N_{\rho,\text{b}}$,
    \Statex Relative 1D offsets for the interior $r_{\text{i},k} \gets Wy+x$ for each ($k$-th) point $(y, x) \in N_{\rho,\text{i}}$,
    \Statex Limit for the count of detected markers $L_m$,
    \Statex Limit for the count of detected sun points $L_s$,
    \Statex Termination pixel sequence $E$
\Part{Variables} \Comment{Stored statically in the RAM}
    \Statex Camera image array $I[WH]$, \Comment{Flattened into 1D}
    \Statex Array of the detected markers $M$ of size $L_m$,
    \Statex Count of the detected markers $c_m$, 
    \Statex Array of the detected sun points $S$ of size $L_s$,
    \Statex Count of the detected sun points $c_s$
\alghline
\Proc{CPUMarkerDetection}{}
    \State $c_m, c_s \gets 0$ \Comment{Reset the counts of the detected points}
    \State $i \gets O_c$ \Comment{Initial offset for the pixel address}
    \State $I[WH-1] \gets E$ \Comment{Write the termination to the end}
\Label{LOOP}
    \If{$I[i+O_c] = E$}
        \textbf{return} \Comment{Quit if terminated}
    \EndIf
    \State $i \gets i + 1$ \Comment{Increment the pixel address}
    \If{$I[i] \leq T_m$}
        \textbf{goto} \texttt{LOOP} \Comment{Central pixel intensity test}
    \EndIf
    \If{$I[i] - I[i+r_{b,1}] \leq T_d$} \Comment{Decide the next test}
        \If{$I[i] \geq T_s$}
            \textbf{goto} \texttt{SUN TEST} 
        \Else 
            \textbf{ goto} \texttt{LOOP}
        \EndIf
    \Else
        \textbf{ goto} \texttt{MARKER TEST}
    \EndIf
\Label{SUN TEST} \Comment{When the first boundary pixel is bright enough}
    \If{$c_s = L_s$} \Comment{Check the sun points count limit}
        \State $I[i+O_c] \gets E$ \Comment{Add the termination if reached}
        \State \textbf{goto} \texttt{LOOP}
    \EndIf
    \ForUnroll{$k \in [2, ..., \len{N_{\rho,\text{b}}}]$} \Comment{The remaining boundary}
        \If{$I[i] - I[i+r_{\text{b},k}] > T_d$}
            \textbf{goto} \texttt{LOOP}
        \EndIf
    \EndFor
    \ForUnroll{$k \in [1, ..., \len{N_{\rho,\text{i}}}]$} \Comment{The whole interior}
        \State $I[i+r_{\text{i},k}] \gets 0$ \Comment{Clear the interior pixel}
    \EndFor
    \State $S[c_s] \gets i$ \Comment{Store the pixel address as a sun point}
    \State $c_s \gets c_s + 1$ \Comment{Increment the count of the detected sun points}
    \State \textbf{goto} \texttt{LOOP}
\Label{MARKER TEST} \Comment{When the first boundary pixel is dark enough}
    \ForUnroll{$k \in [2, ..., \len{N_{\rho,\text{b}}}]$} \Comment{The remaining boundary}
        \If{$I[i] - I[i+r_{\text{b},k}] \leq T_d$}
            \textbf{goto} \texttt{LOOP}
        \EndIf
    \EndFor
    \State $p \gets 0$; $i_p \gets i$ \Comment{Peak pixel value and address}
    \ForUnroll{$k \in [1, ..., \len{N_{\rho,\text{i}}}]$} \Comment{The whole interior}
        \If{$I[i+r_{\text{i},k}] > p$} \Comment{Find and store the peak}
            \State $p \gets I[i+r_{\text{i},k}]$; $i_p \gets i + r_{\text{i},k}$
        \EndIf
        \State $I[i+r_{\text{i},k}] \gets 0$ \Comment{Clear the interior pixel}
    \EndFor
    \State $M[c_m] \gets i_p$ \Comment{Store the peak pixel address as a marker}
    \State $c_m \gets c_m + 1$ \Comment{Increment the count of the detected markers}
    \If{$c_m = L_m$} \Comment{Check the markers count limit}
        \State $I[i+O_c] \gets E$ \Comment{Add the termination if reached}
    \EndIf
    \State \textbf{goto} \texttt{LOOP}
\end{algorithmic}
\end{algorithm}

All the loops are unrolled before compilation, which handles the cases of incorrect compiler optimisation. The termination pixel sequence $E$ is written to the image when either of the counts $c_m$ or $c_s$ reaches its limit, in order to reduce the processing time of excessively noisy or bright frames.

{The time complexity of the proposed FIMD-CPU approach in Alg. \ref{alg:cpu} is $O(W H \rho)$. This does not differ from the time complexity of the unoptimised solution, as all frame pixels still need to be tested for the potential of being a marker or a sun point using the circular boundary. The number of points on the boundary is linearly dependent on the radius $\rho$, approximately as $4\lfloor \sqrt{2}\rho \rfloor$. The substantial performance improvement is achieved through the minimisation of required processor instructions, as the CPU solution can terminate the segment test early if it fails for any of the tested points on the circular boundary. Additionally, significant acceleration is provided by hardcoding most of the constants instead of using variables in RAM memory, thereby reducing memory accesses to only those required for reading and writing frame pixels and storing detection results. Thus, the memory requirements of the proposed approach are directly determined by the sizes of the variables specified in the algorithm.}

\subsection{{FIMD-GPU:} Equivalent pixel-wise approach using GPUs}

\begin{figure}[b!]
  \centering
  \includegraphics[width=\linewidth]{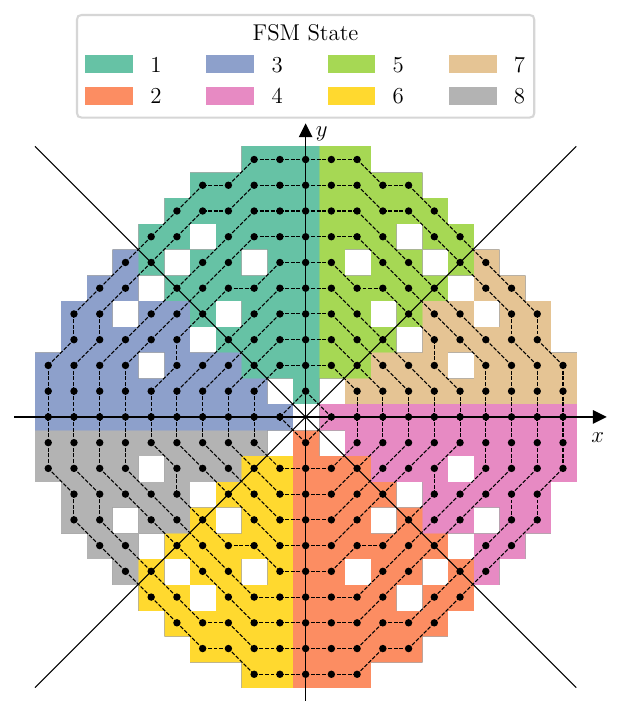}
  \caption{Bresenham's circle approximations for radii $\rho=1,...,10$ with pixels coloured by the corresponding generating FSM states. {The white squares represent points that don’t belong to any Bresenham’s circle approximation. Based on numerical observations for $\rho \in [1, 10^6]$, the number of the points converges to approx. 9.97 \% of all points in the grid ($\mathbb{Z}^2$).}}
  \label{fig:fsm_octants}
\end{figure}

{GPU-based feature and blob detectors discussed in the literature typically rely on computation APIs such as OpenCL or CUDA \parencite{Acharya2014, Oro2016, Li2023}. In contrast, the proposed approach is implemented using standard OpenGL Compute Shaders. Although OpenCL and CUDA libraries provide better floating point precision and more extensive control over GPU resources, they are restricted to GPUs that support these functionalities. The proposed FIMD-GPU approach offers improved portability to embedded systems with low-end CPUs and integrated GPUs that support at least version 3.1 of the OpenGL for Embedded Systems (ES) API. The OpenGL API is sufficient because the parallel computations are based solely on integer arithmetic operations. Furthermore, no additional software support beyond GPU drivers is required for the proposed approach.}

The proposed approach uses a simple procedure to generate points belonging to the Bresenham's square-lattice approximation of an octant of a circle with an integer radius $\rho$. The Alg. \ref{alg:bresenham_fsm} is essentially a finite-state machine (FSM) exploiting the symmetry between the octants of the circle. {The FSM is initialised using the only parameter $\rho$. Per a single call, the proposed procedure produces a single point on the circular boundary. This is followed by an update of the internal state of the FSM, which consists of the FSM state $s$, octant coordinates $x$ and $y$, and an auxiliary decision parameter $P$. To obtain correct points, the radius $\rho$ must not change between consecutive FSM procedure calls.} The FSM states are numbered from 0 (internal update state) to 9 (termination state). The points generated by the individual FSM states 1 to 8 are illustrated in the Fig. \ref{fig:fsm_octants}.

\begin{algorithm}[t!]
 \fontfamily{ptm}\fontsize{8.5}{10.5}\selectfont
\caption{A FSM for generation of Bresenham's circles.}
\label{alg:bresenham_fsm}
\begin{algorithmic}[1]
\Part{FSM Initialisation}
    \Statex $s \gets 1$, \Comment{Initial state of the FSM} 
    \Statex $x, y \gets 0, \rho$, \Comment{Initial relative coordinates} 
    \Statex $P \gets 3-2\rho$ \Comment{Auxiliary decision parameter}
\alghline
\Proc{GetNextBresenhamPoint}{ }
    \If{$s = 0$} \Comment{Update the variables (internal state)}
        \State $x \gets x + 1$
        \If{$P < 0$}
            $P \gets P + 4x + 6$
        \Else
            $\; y \gets y - 1$,
            $P \gets P + 4(x-y) + 10$
        \EndIf
        \If{$x \leq y$}
            $s \gets 1$
        \textbf{else}
            $\; s \gets 9$
        \EndIf
    \EndIf
    \If{$s = 1$}
        $s \gets 2$,
        \textbf{return} $+y, -x$ \Comment{$[90^\circ, 135^\circ]$}  
    \ElsIf{$s = 2$}
        \If{$x < y$}
            $s \gets 3$
        \textbf{else if}{$\; x > 0$}\textbf{ then}
            $s \gets 5$
        \textbf{else}
            $\; s \gets 0$
        \EndIf
        \State \textbf{return} $-y, +x$  \Comment{$[270^\circ, 315^\circ]$}
    \ElsIf{$s = 3$}
        $s \gets 4$,
        \textbf{return} $+x, -y$  \Comment{$(135^\circ, 180^\circ]$}
    \ElsIf{$s = 4$} 
        \If{$x > 0$}
            $s \gets 5$
        \textbf{else}
            $\; s \gets 0$
        \EndIf
        \State \textbf{return} $-x, +y$ \Comment{$(315^\circ, 360^\circ]$}
    \ElsIf{$s = 5$}
        $s \gets 6$,
        \textbf{return} $+y, +x$  \Comment{$[45^\circ, 90^\circ)$}
    \ElsIf{$s = 6$}
        \If{$x < y$}
            $s \gets 7$
        \textbf{else}
            $\; s \gets 0$
        \EndIf
        \State \textbf{return} $-y, -x$  \Comment{$[225^\circ, 270^\circ)$}
    \ElsIf{$s = 7$}
        $s \gets 8$,
        \textbf{return} $+x, +y$  \Comment{$(0^\circ, 45^\circ)$}
    \ElsIf{$s = 8$}
        $s \gets 0$,
        \textbf{return} $-x, -y$  \Comment{$(180^\circ, 225^\circ)$}
    \Else
        \textbf{ return} \textit{Invalid point} \Comment{Termination state}
    \EndIf
\end{algorithmic}
\end{algorithm}

The parallel marker detection, which uses the aforementioned FSM algorithm, is described in more detail in Alg. \ref{alg:gpu}. The circular boundary pixel tests for the marker or sun point potentials are split into two steps. In the first step, the minimum and maximum of all values of the boundary pixels are determined to be used for the potentials evaluation using the thresholds in the second step. This approach takes advantage of the fact that the relation \eqref{eq:sun_pt_potential_min} is equivalent to the case when the central pixel is a sun point, and the relation \eqref{eq:marker_potential_max} similarly represents the case of a marker.

\begin{equation}
\label{eq:sun_pt_potential_min}
    \begin{split}
        &\forall\, x, y: I[r,c] - I[r+y,c+x] < T_d\\
        &\Leftrightarrow \max_{x,y}(I[r,c] - I[r+y,c+x]) < T_d\\
        &\Leftrightarrow I[r,c] - \min_{x,y}(I[r+y,c+x]) < T_d 
    \end{split}
\end{equation}

\begin{equation}
\label{eq:marker_potential_max}
    \begin{split}
        &\forall\, x, y: I[r,c] - I[r+y,c+x] \geq T_d\\
        &\Leftrightarrow \min_{x,y}(I[r,c] - I[r+y,c+x]) \geq T_d\\
        &\Leftrightarrow I[r,c] - \max_{x,y}(I[r+y,c+x]) \geq T_d 
    \end{split}
\end{equation}

{The boundary pixel summarisation, which uses the intermediate variables $b_\text{min}$ and $b_\text{max}$, requires more comparisons than the proposed CPU solution, as the segment tests must be completed entirely to obtain the intermediate results. Therefore, this summarisation is suitable only for parallel processing and is also utilised similarly in a FIMD-FPGA approach presented later on.}

\begin{algorithm}[b!]
 \fontfamily{ptm}\fontsize{8.5}{10.5}\selectfont
\caption{The proposed approach using a GPU shader.}
\label{alg:gpu}
\begin{algorithmic}[1]
\Part{Constants}
    \Statex Limit for the count of detected markers $L_m$,
    \Statex Limit for the count of detected sun points $L_s$
\Part{Variables}
    \Statex Camera image array $I[H,W]$, \Comment{2D, row-major order, 8-bit}
    \Statex Array of the detected markers $M$ of size $L_m$, \Comment{Buffer}
    \Statex Count of the detected markers $c_m$, \Comment{Atomic counter}
    \Statex Array of the detected sun points $S$ of size $L_s$, \Comment{Buffer}
    \Statex Count of the detected sun points $c_s$, \Comment{Atomic counter}
    \Statex Pixel coordinates in the image $r$, $c$,
    \Statex Current min. and max. boundary pixel values $b_{\text{min}}$, $b_{\text{max}}$
\alghline
\Proc{ProcessPixel}{$r$, $c$} \Comment{Compute Shader Program}
    \If{$c_m \geq L_m$ \textbf{or} $c_s \geq L_s$}
        \textbf{return}
    \EndIf
    \If{$I[r, c] \geq T_m$}
        \State $b_{\text{min}}, b_{\text{max}} \gets 255, 0$
        \While{$x, y \gets \Call{GetNextBresenhamPoint}{ }$}
            \If{$I[r+y, c+x] > b_{\text{max}}$}
                $b_{\text{max}} \gets I[r+y, c+x]$
            \EndIf
            \If{$I[r+y, c+x] < b_{\text{min}}$}
                $b_{\text{min}} \gets I[r+y, c+x]$
            \EndIf
        \EndWhile
        \If{$I[r,c] - b_{\text{max}} \geq T_d$}
            \State $M[c_m \gets c_m + 1] \gets r, c$
        \Else
            \If{$I[r,c] \geq T_s$ \textbf{and} $I[r,c] - b_{\text{min}} < T_d$}
                \State $S[c_s \gets c_s + 1] \gets r, c$
            \Else
                \State \textbf{return} 
            \EndIf
        \EndIf
    \EndIf
\end{algorithmic}
\end{algorithm}

Due to concurrency of the computations, the results stored in the arrays are generally unordered and must be sorted first in order to make use of the subsequent filtering. Additionally, the parallel approach can not easily utilise the neighbourhood interior clearing method because of the concurrent memory access. Thus, the concentrated results have to be represented by, e.g., the average coordinates (centres of mass) of the clusters. 

{The time complexity of the proposed FIMD-GPU approach in Alg. \ref{alg:gpu} can be assumed to be $O(\rho)$ due to the parallel computation units processing each of the $WH$ pixels independently. The number of points generated by the FSM procedure in Alg. \ref{alg:bresenham_fsm} grows linearly with radius, similar to the FIMD-CPU approach, and updates of the FSM states have a complexity of $O(1)$. Video memory usage, similar to the FIMD-CPU, is primarily determined by frame resolution, radius, and sizes of arrays for detection results.}

\subsection{{FIMD-FPGA: Equivalent streaming FPGA architecture}}

\begin{algorithm}[b!]
 \fontfamily{ptm}\fontsize{8.5}{10.5}\selectfont
\caption{The proposed streaming FPGA architecture.}
\label{alg:fpga}
\begin{algorithmic}[1]
\Part{Signals} \Comment{Global (parallel) variables}
    \Statex Pixel coordinates in the image $r$, $c$,
    \Statex Consecutive rows array $R[2\rho+1,W]$, \Comment{RAM}
    \Statex Central pixel values array $P[2\rho+1]$, \Comment{LUT}
    \Statex Boundary maxima values array $B_{\text{max}}[2\rho+1]$, \Comment{LUT}
    \Statex Boundary minima values array $B_{\text{min}}[2\rho+1]$, \Comment{LUT}
\Part{Variables} \Comment{Local (sequential) variables}
    \Statex Current pixel indices $s_r, s_c \in \{0, ..., 2\rho\}$,
    \Statex Auxiliary indices $i_r, i_c \in \{0, ..., 2\rho\}$,
    \Statex Relative circle coordinates $x, y \in \{-\rho, ..., \rho\}$
\alghline
\Proc{ProcessSegment}{$r$, $c$} \Comment{Synchronous process}
    \If{rising edge of the valid pixel clock $f_{\text{pix}}$}
        \State $s_r, s_c \gets r, c \text{ mod } (2\rho+1)$
        \State $i_r, i_c \gets (s_r + \rho), (s_c + \rho) \text{ mod } (2\rho+1)$
        \State $P[i_c] \gets R[i_r, c]$ \Comment{Store the central value}
        \While{$x, y \gets \Call{GetNextBresenhamPoint}{ }$}
            \State $i_r, i_c \gets (s_r + \rho + y), (s_c + \rho - x) \text{ mod } (2\rho+1)$
            \If{$R[i_r, c] > B_{\text{max}}[i_c]$}
                $B_{\text{max}}[i_c] \gets R[i_r, c]$
            \EndIf
            \If{$R[i_r, c] < B_{\text{min}}[i_c]$}
                $B_{\text{min}}[i_c] \gets R[i_r, c]$
            \EndIf
        \EndWhile
        \If{$r \geq 2\rho$ \textbf{and} $c \geq 2\rho$} \Comment{The processing delay}
            \If{$P[s_c] > T_m$}
                \If{$P[s_c] - B_{\text{max}}[s_c] \geq T_d$}
                    \State \textbf{output} marker at $r - \rho$, $c - \rho$  
                \ElsIf{$P[s_c] > T_s$ \textbf{and} $P[s_c] - B_{\text{min}}[s_c] < T_d$}
                    \State \textbf{output} sun point at $r - \rho$, $c - \rho$  
                \EndIf
            \EndIf
            \State $B_{\text{min}}[s_c], B_{\text{max}}[s_c] \gets 255, 0$ \Comment{Reset the array values}
        \EndIf
        
    \EndIf
\end{algorithmic}
\end{algorithm}

The main idea behind the proposed architecture for FPGAs is that the markers and the sun points can be detected synchronously with the original pixel frequency ($f_{\text{pix}}$) of the CMOS imaging sensor. To make FPGAs capable of direct pixel stream evaluation, all of the prior pixels necessary for the detection have to be buffered in the memory. Conceptually similar approaches, {which leverage the pipelining technique to process multiple data elements concurrently as well,} are commonly present in the literature \parencite{Huang2018, Haijun2010, Heo2013, Cizek2018}, {including several general approaches to blob detection proposed for FPGAs \parencite{AcevedoAvila2016, Petrovic2017}.}

The proposed architecture for the isolated marker detection is outlined by Alg. \ref{alg:fpga}. The pixel array $R$ represents $2\rho+1$ row arrays of $W$ pixels, showing that the proposed approach does not require storing all frame pixels in order to provide all of the required information for the following algorithms in the localisation system pipeline. The auxiliary arrays used by the proposed architecture are conveniently filled in a circular FIFO manner by the parallel camera interface (DCMI) of a CMOS sensor, which produces one pixel at every rising edge of a valid pixel clock (i.e., when the VSYNC and HSYNC interface signals are active). The interface outputs pixels in a row-major order. This causes a minimal processing delay until $2\rho(W+1)$ pixels are buffered, which is required for the completion of the first square image segment with a size of $2\rho+1$ pixels containing a circle with the radius $\rho$.

At every valid pixel clock rising edge, the procedure reads the buffered rows from $R$ in parallel. It then processes a single column finished by the currently streamed pixel in order to evaluate all of the circular neighbourhoods in overlapping frame segments containing that column. The array $B_{\text{min}}$ is initialised with the highest values $\texttt{0xFF}$. All other arrays are initialised with zeros. The arrays are used to store the boundary maxima and minima values, the central pixel values of the segments, and the detected markers and sun points. The arrays are situated either in the RAM or wired directly into the fabric.

The arrays are accessed using auxiliary indices $i_r$, $i_c$, which are obtained from the FSM-generated coordinates relative to the centre of the segment. This approach exploits the symmetry of the Bresenham's circles to update the values in arrays related to multiple overlapping segments, thus it enables parallel evaluation of multiple circles.

The pseudocode in the Alg. \ref{alg:fpga} is similar to the parallel GPU solution described in the Alg. \ref{alg:gpu}, distinguished by its dependence on the sequential pixel stream and by the reduced requirements for the memory resources. {Additionally, unlike the FIMD-GPU approach, which can be easily repeated using multiple radii, the proposed FPGA solution allows for parallel detection across multiple radii by implementing the synchronous process multiple times, incurring only the cost of additional FPGA resources.} As in the GPU solution, the detected points must be approximated using the cluster centres to be filtered out by the sun-marker distance. 

{To summarise utilisation of the FPGA resources, the proposed FIMD-FPGA approach necessitates $(2\rho+1)\cdot W$ bytes of block RAM for storing the number of frame rows based on the selected radius, while all other arrays are limited to a size of $2\rho+1$ bytes and are directly wired using the FPGA fabric signals. The final implementation of the synchronous process in the Alg. \ref{alg:fpga} used 1472 out of 41910 available Adaptive Logic Modules (ALMs).}

\section{Experiments}
\label{sec:experiments}

{This paper focuses on the processing speed of camera data and the potential for real-time deployment using less powerful drone platforms. Our motivation stems from the application of the UVDAR system for testing swarm behaviour in real environments \parencite{Dmytruk2021, Walter2020, Horyna2024, Petracek2020}.}

\subsection{Dataset for evaluation}

The evaluation of the {unoptimised} approach and the proposed approach was performed using a dataset recorded during real-world UAV experiments. {The real-world experiments were carried out in a desert environment at noon with a swarm of 9 UAVs, which maintained mutual distances between 2 and 15 metres. Therefore, the dataset is suitable for demonstration of robustness of the proposed approach to high levels of exterior illumination. A representative sample is shown in Fig. \ref{fig:detector_example}a.} The dataset consists of five raw recordings taken at 60 FPS and with an exposure interval of 500 \textmu s, containing 83914 8-bit image samples in total.

\begin{figure}[b!]
\centering
\begin{subfigure}[b]{.30\linewidth}
\includegraphics[width=\linewidth]{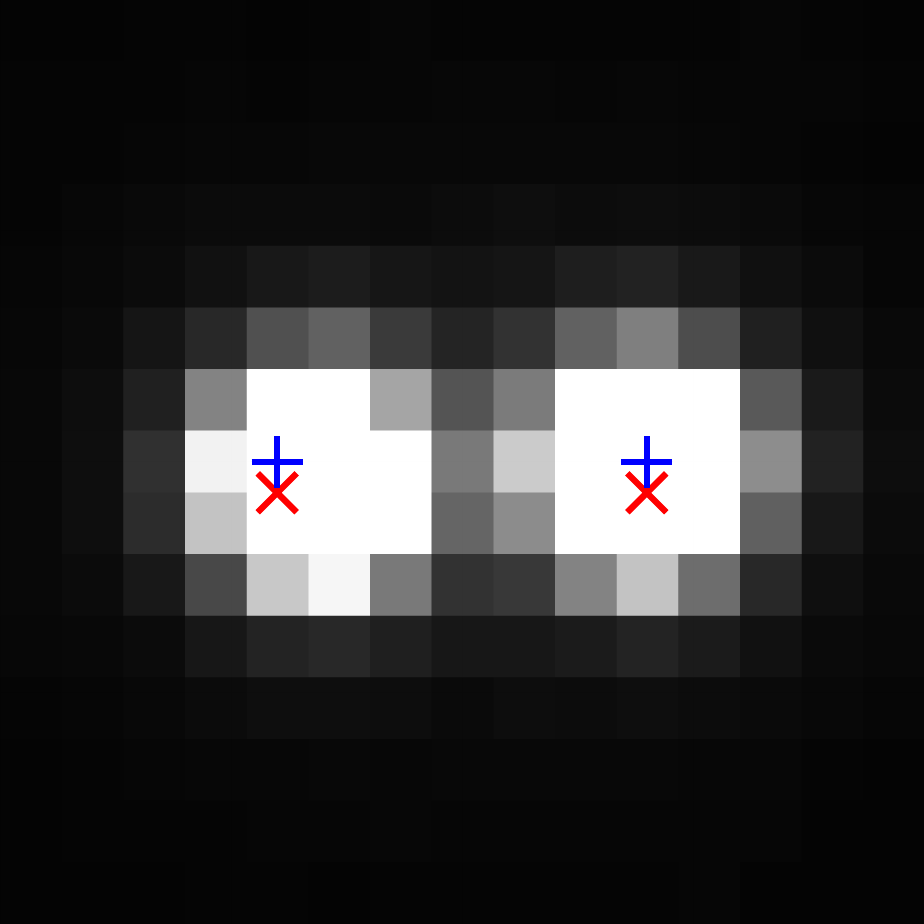}
\caption{\fontsize{8.5}{10}\selectfont{Dist. 1 m.}}\label{fig:marker_dist1m}
\end{subfigure}
\begin{subfigure}[b]{.30\linewidth}
\includegraphics[width=\linewidth]{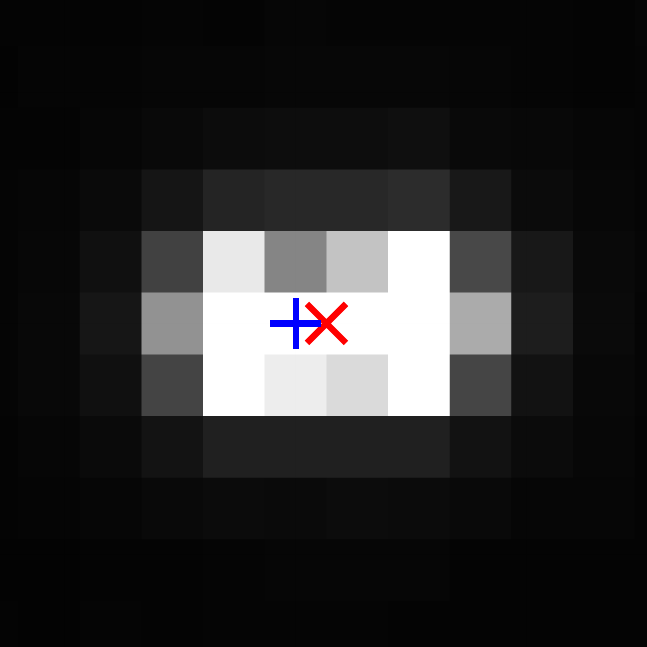}
\caption{\fontsize{8.5}{10}\selectfont{Dist. 2 m.}}\label{fig:marker_dist2m}
\end{subfigure}
\begin{subfigure}[b]{.30\linewidth}
\includegraphics[width=\linewidth]{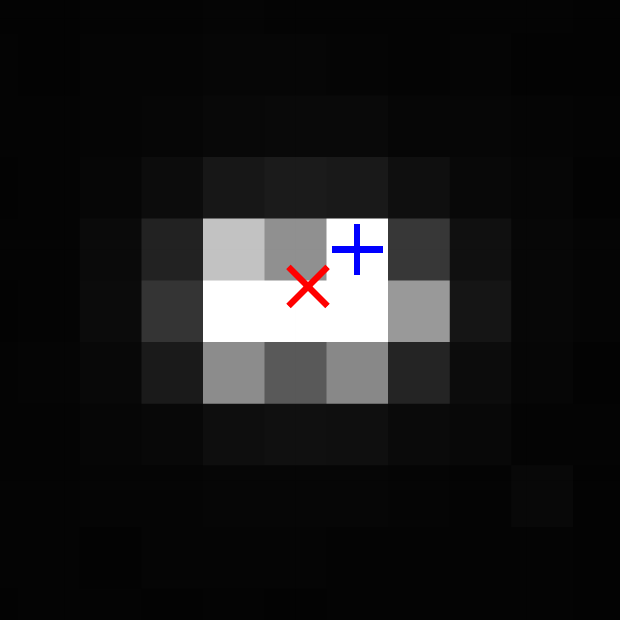}
\caption{\fontsize{8.5}{10}\selectfont{Dist. 3 m.}}\label{fig:marker_dist3m}
\end{subfigure}

\begin{subfigure}[b]{.30\linewidth}
\includegraphics[width=\linewidth]{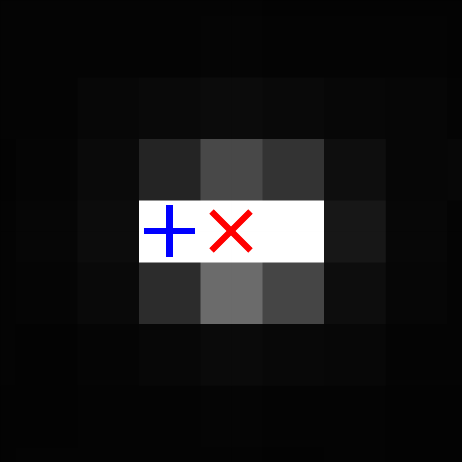}
\caption{\fontsize{8.5}{10}\selectfont{Dist. 5 m.}}\label{fig:marker_dist5m}
\end{subfigure}
\begin{subfigure}[b]{.30\linewidth}
\includegraphics[width=\linewidth]{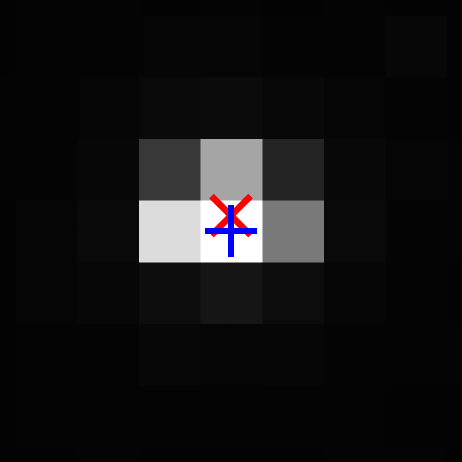}
\caption{\fontsize{8.5}{10}\selectfont{Dist. 7 m.}}\label{fig:marker_dist7m}
\end{subfigure}
\begin{subfigure}[b]{.30\linewidth}
\includegraphics[width=\linewidth]{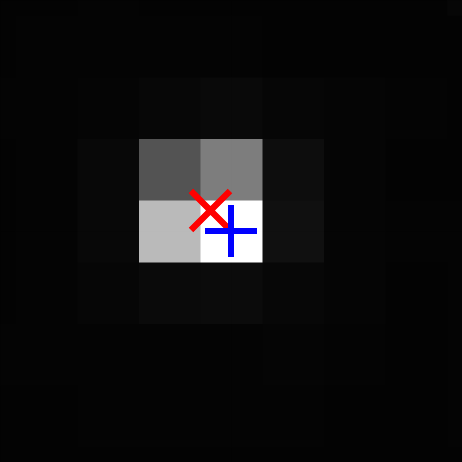}
\caption{\fontsize{8.5}{10}\selectfont{Dist. 9 m.}}\label{fig:marker_dist9m}
\end{subfigure}

\begin{subfigure}[b]{.30\linewidth}
\includegraphics[width=\linewidth]{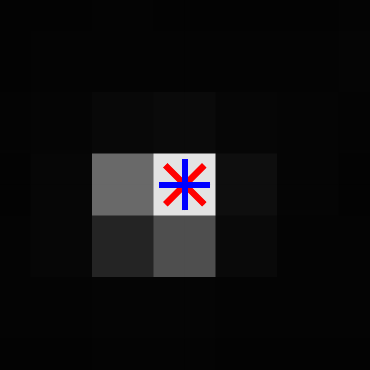}
\caption{\fontsize{8.5}{10}\selectfont{Dist. 11 m.}}\label{fig:marker_dist11m}
\end{subfigure}
\begin{subfigure}[b]{.30\linewidth}
\includegraphics[width=\linewidth]{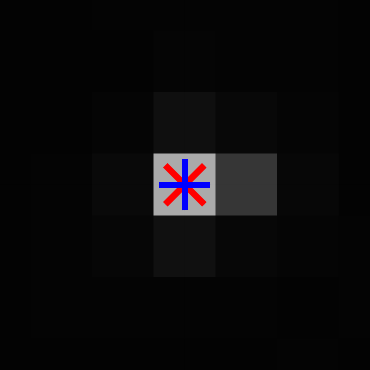}
\caption{\fontsize{8.5}{10}\selectfont{Dist. 15 m.}}\label{fig:marker_dist15m}
\end{subfigure}
\begin{subfigure}[b]{.30\linewidth}
\includegraphics[width=\linewidth]{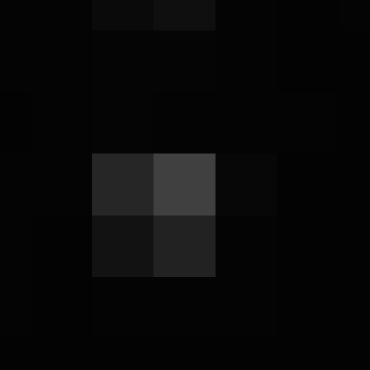}
\caption{\fontsize{8.5}{10}\selectfont{Dist. 20 m.}}\label{fig:marker_dist20m}
\end{subfigure}

\caption{{Detailed examples of isolated UV LED markers at various distances from the camera. Crosses indicate the accuracy of the FIMD approach with parameters $r=3$, $T_m=120$ and $T_d=60$. Blue crosses represent detections from the UIMD/FIMD-CPU approach, red crosses denote centres of clusters of detected markers from the FIMD-GPU/-FPGA approach.}}
\label{fig:markers_examples}
\end{figure}

\begin{table*}[t!]
\centering
 \fontfamily{ptm}\fontsize{7.5}{10.5}\selectfont
\def\arraystretch{1.1}
\begin{tabular}{||@{\hspace{3pt}}l@{\hspace{3pt}}|@{\hspace{3pt}}l@{\hspace{3pt}}|@{\hspace{3pt}}l@{\hspace{3pt}}|@{\hspace{3pt}}l@{\hspace{3pt}}|@{\hspace{3pt}}l@{\hspace{3pt}}||}
\hline
\textbf{Board} & \textbf{System-on-Chip (SoC)} & \textbf{Architecture} & \textbf{Composition of cores, sel. freq.} & \textbf{Additional notes} \\
\hline\hline
Intel NUC 10 & Intel Core i7-10710U & 64-bit x86 & 6x Comet Lake-U @ 1601 MHz & Intel UHD Graphics 630 @ 1150 MHz \\ 
\hline
Raspberry Pi 4B & Broadcom BCM2711 & 64-bit ARM & 4x Cortex-A72 @ 1500 MHz  & Broadcom VideoCore VI @ 500 MHz \\ 
\hline
Raspberry Pi 5 & Broadcom BCM2712 & 64-bit ARM & 4x Cortex-A76 @ 2400 MHz & Broadcom VideoCore VII @ 910 MHz  \\ 
\hline
Terasic DE10-Nano & Intel Cyclone 5CSEBA6U23I7 & 64-bit ARM & 2x Cortex-A9 @ 800 MHz & FPGA 110k LEs @ 26.67 MHz (= $f_{\text{pix}}$) \\ 
\hline
ST Nucleo-H745ZI-Q & ST STM32H745ZIT6 & 32-bit ARM & 1x Cortex-M7 @ 480 MHz & (also contains Cortex-M4 @ 240 MHz) \\ 
\hline
\end{tabular}
\caption{An overview of the hardware platforms selected for evaluation.}
\label{tab:hw_boards}
\end{table*}

{To illustrate the dataset further, selected examples of isolated markers at various distances are shown in Fig. \ref{fig:markers_examples}. The examples show details of projections of UV LED pairs (Fig. \ref{fig:UV_components}a) that have both LEDs directed with an approx. 45$^\circ$ deviation from the optical axis of the camera lens. In Fig. \ref{fig:markers_examples}, the markers were reliably detectable at distances ranging from 2 to 15 meters using detection radii of $\rho=3$ or $\rho=4$. Central pixels remained saturated up to a distance of 9 meters. Depending on the background illumination levels, markers located further away often became indistinguishable from neighbouring pixels in the dataset frames. The sensitivity from the perspective of relative localisation is discussed in more detail in the original UVDAR work \parencite{Walter2018_1}.}

{Additionally, the Fig. \ref{fig:markers_examples} illustrates the accuracy of the proposed FIMD approach. The first encountered interior peaks are stored as the detected marker pixels (blue crosses) by the FIMD-CPU approach. However, the interior peak search is not performed by the FIMD-GPU and the FIMD-FPGA approaches, and all pixels that successfully pass the boundary segment test are stored. Therefore, the detected pixels must be clustered first, and the centers of the clusters are used as the final detected markers (red crosses). An implicit limit on inaccuracy is imposed by the small size of the segment test window, which restricts the number of pixels by which the detection can be off. However, if the FIMD approach were initialized with $r=4$ instead of $r=3$, the closest pair of markers in Fig. \ref{fig:markers_examples}a, which should be detected as a single marker, would be rejected. Thus, it is favourable for the detection process to be repeated for multiple radii and possibly also for multiple threshold levels. However, for the purpose of the performance evaluation presented in this work, only a single radius and a single set of threshold values were assumed.}

\subsection{Hardware overview}

The camera modules originally used for compiling the dataset are the Matrix Vision mvBlueFOX-200wG. These modules are based on the MT9V034 global-shutter monochromatic CMOS imaging sensors, which offer a quantum efficiency of around 37\% at the wavelength of 395 nm, and a resolution of 752x480 pixels at frame rates up to 93 FPS in a continuous operation mode. However, the maximal achievable frame rate is limited down to about 60 FPS when the USB 2.0 interface is used. To overcome this limitation, a custom camera module with the same CMOS sensor was designed in order to directly expose the 10-bit 26.67 MHz parallel camera interface to the MCU and FPGA development boards.

The computational hardware platforms used for evaluation of the proposed solutions are listed in Tab. \ref{tab:hw_boards}. The main chipset used by the MRS drone platform \parencite{Hert2023} is the Intel Core i7-10710U. It runs the UVDAR system as a ROS package under the Linux-based MRS UAV system \parencite{Baca2021}. To perform the evaluation in an embedded chipset environment, the popular single-board computer boards Raspberry Pi 4B and Raspberry Pi 5 were selected. The MCU Nucleo development board was used for a demonstration of an effective deployment of the optimised CPU approach on a single 32-bit ARM Cortex-M7 core, running at just 480 MHz. For verification of the proposed FPGA architecture, the board Terasic DE10-Nano incorporating the Intel Cyclone V SoC was chosen.

\subsection{{Dataset evaluation}}

\begin{figure}[b!]
  \centering
  \includegraphics[width=\linewidth]{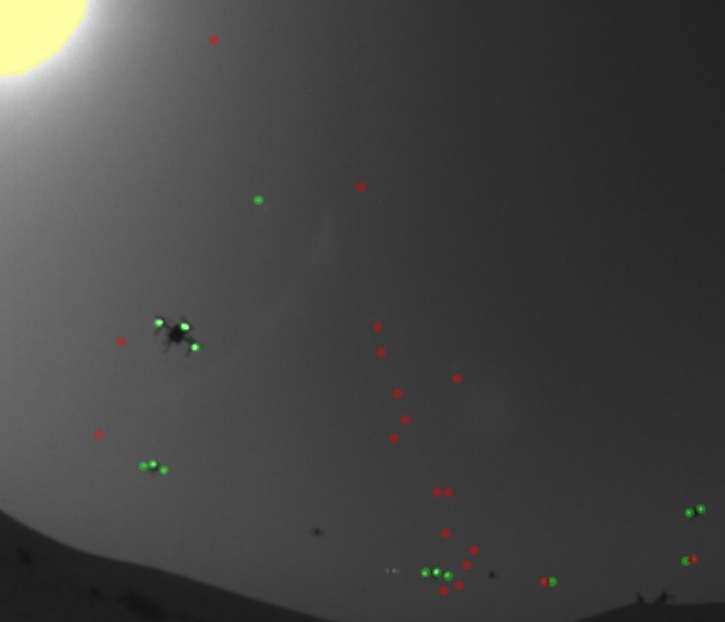}
  \caption{{Sample frame with indicated detection results. Sun pixels detected by FIMD approach are shown in yellow. Isolated markers detected by both FIMD and SBD approaches are shown in green. The SBD approach detects additional blobs, highlighted in red, with only the two rightmost blobs being true positives. The topmost detection, common to both approaches, is caused by a lens flare, thus it is a false positive.}}
  \label{fig:dataset_fimd_sbd_sample}
\end{figure}

The {UIMD} approach to the marker detection was compared with the proposed FIMD approach using the dataset of raw camera frames and the hardware listed in Tab. \ref{tab:hw_boards}. {Additionally, an existing Simple Blob Detection (SBD) approach from the OpenCV library was selected for comparison with the proposed FIMD approach.}

{The SBD approach starts by applying multi-level thresholding to a greyscale image, using threshold levels that range linearly from $T_\text{min}$ to $T_\text{max}$ with an incremental step of $T_\text{step}$. This process produces binary images, from which contours of connected components are extracted. The final step involves filtering out contours based on parameters such as color, area, circularity, or convexity. For a fair comparison with the FIMD approach on the dataset, all thresholds were set to $T_\text{min}=T_m=60$, $T_\text{max}=T_s=240$, and $T_\text{step}=T_d=30$. Contours were excluded if their color was not white or if their area exceeded $\pi r^2$, where $r$ was the largest selected radius for the FIMD approach. No additional filters were applied.}

\begin{figure}[t!]
  \centering
  \includegraphics[width=\linewidth]{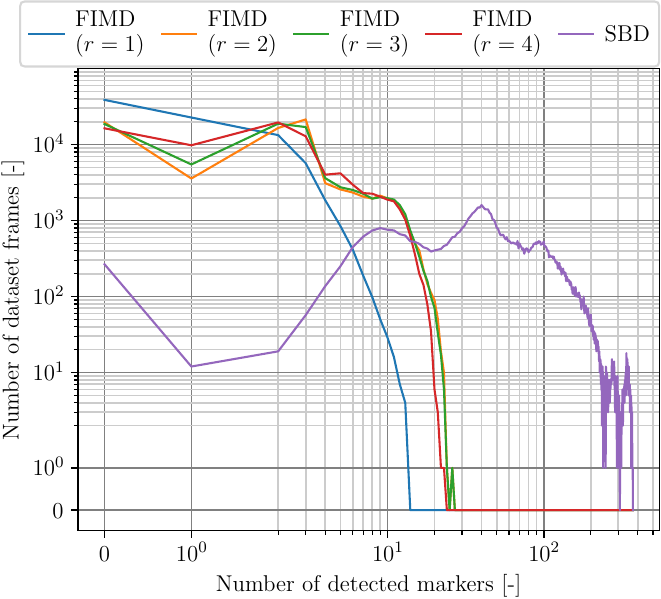}
  \caption{{Histograms of the markers detected by the FIMD and SBD approaches. The radii 1 to 4 correspond to the maximum blob area, which was set to $\pi4^2 \approx 50\; \text{px}^2$. It can be noted that for $2\leq r\leq4$, the counts of FIMD positives are similarly distributed across the dataset up to the maximum of 26 markers. The SBD generates many false positives that do not meet the marker detection criteria defined by the FIMD algorithm. }}
  \label{fig:dataset_hist_markers}
\end{figure}

{The detection results for both approaches are shown in Fig. \ref{fig:dataset_fimd_sbd_sample}, which demonstrates that the SBD approach has similar detection sensitivity but results in more false detections. Fig. \ref{fig:dataset_hist_markers} presents multiple histograms of detected markers by the FIMD approach with multiple radii, and by the SBD approach. With lower $T_\text{step}$ values, the SBD approach achieved finer resolution at the expense of performance due to the production of more binary images. A related issue was an increased occurrence of false detections, mainly caused by small binary artifacts resulting from finer thresholding of bright background areas, especially around the sun projection. In contrast, when the values of $T_\text{min}$ or $T_\text{step}$ increased, the SBD approach could not match the sensitivity of the proposed approach. It should also be noted that detecting the sun using the SBD approach must be performed as a separate task with different filtering parameters; hence, it was omitted from the performance evaluation on the dataset. This fact highlights an indisputable advantage of the FIMD approach.}


{The lower sensitivity of the SBD algorithm can be attributed to its underlying working principle. The SBD approach implements global thresholding in incremental steps across the entire image, disregarding small variations in image intensity between thresholds. In contrast, the proposed FIMD algorithm applies differential thresholding based on the local intensity of the central pixel.}

\begin{figure}[b!]
  \centering
  \includegraphics[width=\linewidth]{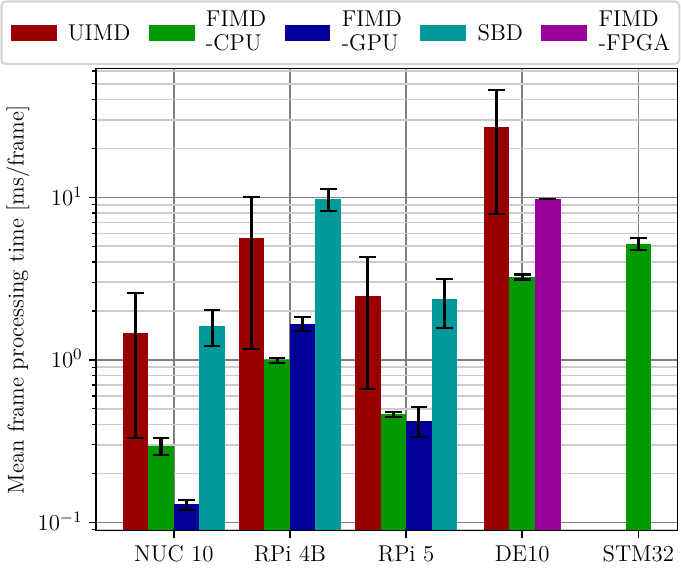}
  \caption{{Mean values of processing times for all dataset frames across all evaluated approaches. Standard deviations are indicated as error ranges of individual bars.}}
  \label{fig:performance_mean_times_bars}

\end{figure}

\begin{table}[t!]
\centering
 \fontfamily{ptm}\fontsize{7.5}{10.5}\selectfont
\def\arraystretch{1.2}
\begin{tabular}{||@{\hspace{3pt}}l@{\hspace{3pt}}|@{\hspace{3pt}}c@{\hspace{3pt}}|@{\hspace{3pt}}c@{\hspace{3pt}}|@{\hspace{3pt}}c@{\hspace{3pt}}|@{\hspace{3pt}}c@{\hspace{3pt}}||}
\hline
\textbf{Board} & \textbf{UIMD} & \textbf{FIMD-CPU} & \textbf{FIMD-GPU} & \textbf{SBD}\\
\hline\hline
NUC 10 & 1.46 $\pm$ 1.13 & 0.29 $\pm$ 0.04 & 0.13 $\pm$ 0.01 & 1.63 $\pm$ 0.40 \\ 
\hline
RPi 4B & 5.63 $\pm$ 4.46 & 0.99 $\pm$ 0.03 & 1.67 $\pm$ 0.16 & 9.76 $\pm$ 1.55 \\ 
\hline
RPi 5 & 2.47 $\pm$ 1.81 & 0.46 $\pm$ 0.02 & 0.42 $\pm$ 0.09 & 2.36 $\pm$ 0.78 \\ 
\hline
DE10 & 26.98 $\pm$ 19.08 & 3.24 $\pm$ 0.11 & N/A & N/A \\ 
\hline
STM32 & N/A & 5.18 $\pm$ 0.44 & N/A & N/A \\ 
\hline
\end{tabular}
\caption{{Mean and standard deviation values (expressed as $\mu \pm \sigma$) of processing times of all dataset frames for all evaluated approaches. For the FIMD-FPGA approach, the mean value was 9.76 ms with a zero deviation.}}
\label{tab:mean_times}
\end{table} 

{To assess the computational efficiency of the CPU-based approaches and of the FIMD-GPU approach, processing times were measured repeatedly, ranging from 100 to 1000 iterations, on a CPU core that was isolated from the Linux kernel.} {All implementations of the approaches were compiled using the highest optimisation level available from the compiler, specifically -O3. This ensured that all implementations were able to run as efficiently as possible.} The CPU and GPU clocks were fixed at the maximum available frequency. {For the FIMD-FPGA approach}, all processing times were equal the total duration of the VSYNC pulse, which is constant for any exposure settings and equal to 9756.5 \textmu s. 

{Mean frame processing times including standard deviations are shown in Fig. \ref{fig:performance_mean_times_bars} and summarised in Tab. \ref{tab:mean_times}. Based on the standard deviation values, it is evident that the FIMD approach offers significantly more consistent performance compared to the UIMD and SBD approaches when processing frames with varying brightness and with differing numbers of contained markers.}

\subsection{{Performance gain comparison}}

\begin{figure}[t!]
  \centering
  \includegraphics[width=\linewidth]{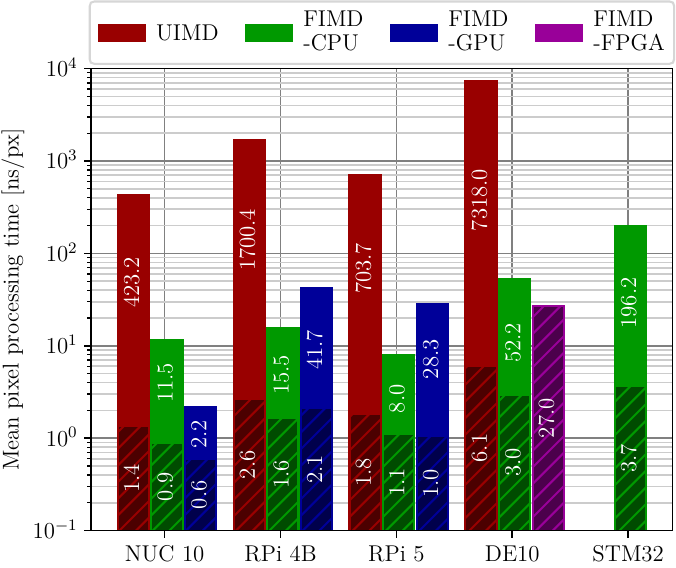}
  \caption{Performance {gain} comparison for all hardware platforms. The approximations of $t_{\text{none}}$ and $t_{\text{det}}$ are visualised, including their values as the darker hatched segments of the bars and the whole bars, respectively.}
  \label{fig:performance_comparison}
\end{figure}

The comparison is shown in Fig. \ref{fig:performance_comparison} and the achieved acceleration factors are summarised in the Tab. \ref{tab:approx_vals_comparison}.The measured mean frame processing times $t_{\text{frame}}$ were approximated by a linear function \eqref{eq:det_lin_approx} of the total number of image pixels $n_{\text{det}}$ successfully passing the neighbourhood segment test:
\begin{equation}
    t_{\text{frame}}(n_{\text{det}}) \approx (WH - n_{\text{det}}) t_{\text{none}} + n_{\text{det}} t_{\text{det}}.
    \label{eq:det_lin_approx}
\end{equation}

The approximation parameters are $t_{\text{det}}$ for the pixels producing either a marker or a sun point, and $t_{\text{none}}$ for the rest of the frame pixels. They represent estimations of processing times per pixel independent on the dataset image resolution. {The reasoning behind this approximation is that the processing times per pixel vary across frames containing markers for all evaluated approaches except the FIMD-FPGA approach. Thus, the approximation given in \eqref{eq:det_lin_approx} allows for a comparison of the performance of all approaches at the pixel level.}

\begin{table}[b!]
 \fontfamily{ptm}\fontsize{7.5}{10.5}\selectfont
\centering
\def\arraystretch{1.2}
\begin{tabular}{||@{\hspace{3pt}}l@{\hspace{3pt}}|l@{\hspace{3pt}}|c|c|c|c||}
\hline
\multirow{2}*{\shortstack{\\ \textbf{Board}\\ \textbf{{(local)}}}} & \multirow{2}*{\textbf{{Approach}}} & \multicolumn{2}{c}{\textbf{{UIMD (NUC 10)}}} & \multicolumn{2}{|c||}{\textbf{{UIMD (on local)}}}\\
\hhline{~~----}
& &\makecell{\fontfamily{ptm}\fontsize{8.0}{10.5}\selectfont vs.\\ $t_{\text{none}}$} & \makecell{\fontfamily{ptm}\fontsize{8.0}{10.5}\selectfont vs.\\ $t_{\text{det}}$} & \makecell{\fontfamily{ptm}\fontsize{8.0}{10.5}\selectfont vs.\\ $t_{\text{none}}$} & \makecell{\fontfamily{ptm}\fontsize{8.0}{10.5}\selectfont vs.\\ $t_{\text{det}}$} \\
\hline\hline

\multirow{2}*{NUC 10} & UIMD & {(1.00)} & {(1.00)} & {(1.00)} & {(1.00)} \\ 
\hhline{~-----}
& {FIMD-}CPU & {\color{Green} 1.56} & {\color{Green} 36.80} & {({\color{Green} 1.56})} & {({\color{Green} 36.80})} \\ 
\hhline{~-----}
 & {FIMD-}GPU & {\color{Green} 2.33} & {\color{Green} 192.36} & {({\color{Green} 2.33})} & {({\color{Green} 192.36})} \\ 
\hline

\multirow{3}*{RPi 4B} & {UIMD} & {\color{Red} 0.54} & {\color{Red} 0.25} & {(1.00)} & {(1.00)}  \\ 
\hhline{~-----}
 & {FIMD-}CPU & {\color{Red} 0.88} & {\color{Green} 27.30} & {\color{Green} 1.62} &  {\color{Green} 109.70}  \\ 
\hhline{~-----}
 & {FIMD-}GPU & {\color{Red} 0.67} & {\color{Green} 10.15} & {\color{Green} 1.24} & {\color{Green} 40.78}  \\ 
\hline

\multirow{3}*{RPi 5} & {UIMD} & {\color{Red} 0.78} & {\color{Red} 0.60} & {(1.00)}  & {(1.00)}  \\ 
\hhline{~-----}
 & {FIMD-}CPU & {\color{Green} 1.27} & {\color{Green} 52.9} & {\color{Green} 1.64} &  {\color{Green} 87.96} \\ 
\hhline{~-----}
 & {FIMD-}GPU & {\color{Green} 1.4} & {\color{Green} 14.95} & {\color{Green} 1.80} & {\color{Green} 24.87}  \\ 
\hline

\multirow{3}*{DE10} & {UIMD} & {\color{Red} 0.23} & {\color{Red} 0.06} & {(1.00)} & {(1.00)}  \\ 
\hhline{~-----}
& {FIMD-}CPU & {\color{Red} 0.47} & {\color{Green} 8.11} & {\color{Green} 2.03}  & {\color{Green} 140.19}   \\ 
\hhline{~-----}
& {FIMD-}FPGA & {\color{Red} 0.05} & {\color{Green} 15.67} & {\color{Red} 0.23}  & {\color{Green} 270.76}   \\ 
\hline

STM32 & {FIMD-}CPU & {\color{Red} 0.38} & {\color{Green} 2.16} & N/A & N/A  \\ 
\hline
\end{tabular}
\caption{Summary of the acceleration factors achieved by the proposed solutions when compared to the {UIMD approach} on the NUC 10 board, and also to the {UIMD approach} on the same local hardware.}
\label{tab:approx_vals_comparison}
\end{table} 


For comparison with the proposed FPGA architecture, the equivalent values of the mean processing times per pixel are shown in the Tab. \ref{tab:approx_vals_comparison} and in Fig. \ref{fig:performance_comparison}. In this case, the values were obtained from the {constant frame processing time} divided by the frame resolution, {because} the proposed approach ensures that $t_{\text{none}} = t_{\text{det}}$, and that an evaluation of a single circle takes exactly $1 / f_{\text{pix}}$ seconds.

\begin{figure}[b!]
  \centering
  \includegraphics[width=\linewidth]{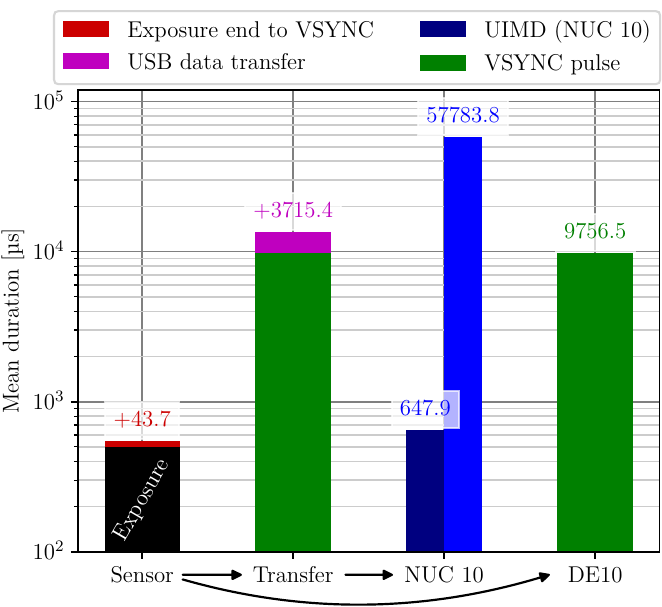}
  \caption{Mean duration of internal sensor delays, data transfers, and processing times for the {UIMD} approach and for the FPGA. Black arrows indicate how the total duration on the two platforms is obtained. The sensor exposure time related to the dataset (black bar) is followed by an additional in-sensor delay (red bar). The mean CPU processing time is shown for all dataset images producing no detection results, and the maximum number of detection results (dark and light blue bars, respectively).}
  \label{fig:sensor_data_pipeline}
\end{figure}

The acceleration factors are not as substantial as in the case of the CPU and GPU. Nonetheless, the main advantage of the proposed architecture is an elimination of the frame transfer via the USB interface, which is required by the other proposed solutions as illustrated in Fig. \ref{fig:sensor_data_pipeline}. The proposed architecture completes the processing of a camera frame when the vertical synchronisation (VSYNC) signal of the CMOS sensor goes to an inactive state. In the case of the other approaches, a host computer is not able to process the frame until the additional USB transfer of the frame completes, which was measured at approx. $3.8$ ms. Thus, the proposed approach minimises the delay of the detection results, measured since the exposure interval of the sensor.

\section{Conclusion}

This paper has proposed a novel fast onboard marker detection strategy that enables visual relative localisation in agile aerial robotic swarms in real-world conditions. For the main objective of the fast detection of rapidly moving isolated LED markers, we have proposed a sequential procedure optimised for CPUs, a parallel compute shader procedure for GPUs, and a streaming FPGA architecture.

The {original unoptimised approach} and the proposed approach were compared on multiple hardware platforms using a real-world dataset of raw recorded camera frames. {To compare the proposed approach with another available baseline, a Simple Blob Detector from the OpenCV library was evaluated on the dataset. The proposed approach outperformed the well-established C++ implementation on all platforms in terms of performance, while maintaining superior detection sensitivity and accuracy.}

On the originally used Intel NUC 10 computer, the proposed approach optimised for CPUs and GPUs offered per-pixel acceleration factors for a successful marker or sun point detection equal to 36.8 and 192.36, respectively, when compared to the state-of-the-art approach on the same hardware. Moreover, the proposed CPU approach enabled real-time processing on a low-end MCU with an average core utilisation around 65\% at the maximum camera frame rate of 93 FPS. In the case of the proposed FPGA architecture, all detection results were available around 3.8 ms in advance of the CPU and GPU frame processing, which represents the maximum achieved overall acceleration.

\section*{Declarations}
\subsection*{Funding}
{This work was funded by the European Union under the project Robotics and advanced industrial production (reg. no. CZ.02.01.01/00/22\_008/0004590), by the Czech Science Foundation (GAČR) under research project no. 23-07517S, and by CTU grant no. SGS23/177/OHK3/3T/13.}

\subsection*{Conflict of interest}
The authors have no conflicts of interest to declare that are relevant to the content of this article.

\subsection*{Ethics approval}
The conducted research is not related to either human or animal use.

\subsection*{Availability of data}
{The implementations of the proposed FIMD approach, along with the dataset of raw camera recordings that support the findings of this study, are publicly accessible online\footnote{\url{https://github.com/ctu-mrs/FIMD}}.}

\subsection*{Author contribution}
V.V. is the author of the entire manuscript. All authors reviewed the manuscript.

\end{document}